\documentclass[11pt,a4paper]{article}
\usepackage[hyperref]{emnlp2018}
\usepackage{times}
\usepackage{latexsym}
\usepackage[justification=centering]{caption}
\usepackage{graphicx}
\usepackage{tabularx}
\usepackage{soul}
\usepackage{algorithm,algorithmic}
\usepackage{textgreek}
\usepackage{multirow}
\usepackage{enumitem}
\usepackage{tabu}
\usepackage{array}
\usepackage{booktabs}
\usepackage{mathtools}
\usepackage{tikz}
\usepackage[normalem]{ulem}
\usepackage{gensymb}
\usepackage{amssymb}
\usepackage{pifont}
\usepackage{tikz}
\usepackage[utf8x]{inputenc}
\setlist{noitemsep}
\usepackage{url}

\aclfinalcopy % Uncomment this line for the final submission
\title{Normalization of Transliterated Words in Code-Mixed Data Using Seq2Seq Model \& Levenshtein Distance}

\author{
        Soumil Mandal, Karthick Nanmaran \\ \\
       	Department of Computer Science \& Engineering \\ 
        SRM Institute of Science \& Technology, Chennai, India\\
        \textcolor{black!50}{\{soumil.mandal, karthicknanmaran\}@gmail.com}                
}

\date{}

\begin{document}
\maketitle

\begin{abstract}
Building tools for code-mixed data is rapidly gaining popularity in the NLP research community as such data is exponentially rising on social media. Working with code-mixed data contains several challenges, especially due to grammatical inconsistencies and spelling variations in addition to all the previous known challenges for social media scenarios. In this article, we present a novel architecture focusing on normalizing phonetic typing variations, which is commonly seen in code-mixed data. One of the main features of our architecture is that in addition to normalizing, it can also be utilized for back-transliteration and word identification in some cases. Our model achieved an accuracy of 90.27\% on the test data. 
\end{abstract}

\section{Introduction}

With rising popularity of social media, the amount of data is rising exponentially. If mined, this data can proof to be useful for various purposes. In countries where the number of bilinguals are high, we see that users tend to switch back and forth between multiple languages, a phenomenon known as code-mixing or code-switching. An interesting case is switching between languages which share different mother scripts. On such occasions, one of the two languages is typed in it's phonetically transliterated form in order to use a common script. Though there are some standard transliteration rules, for example ITRANS~\footnote{https://en.wikipedia.org/wiki/ITRANS}, ISO~\footnote{https://en.wikipedia.org/wiki/ISO\_15919}, but it is extremely difficult and un-realistic for people to follow them while typing. This indeed is the case as we see that identical words are being transliterated differently by different people based on their own phonetic judgment influenced by dialects, location, or sometimes even based on the informality or casualness of the situation. Thus, for creating systems for code-mixed data, post language tagging, normalization of transliterated text is extremely important in order to identify the word and understand it's semantics. This would help a lot in systems like opinion mining, and is actually necessary for tasks like summarization, translation, etc. A normalizing module will also be of immense help while making word embeddings for code-mixed data. \\
\hspace*{0.5cm} In this paper, we present an architecture for automatic normalization of phonetically transliterated words to their standard forms. The language pair we have worked on is Bengali-English (Bn-En), where both are typed in Roman script, thus the Bengali words are in their transliterated form. The canonical or normalized form we have considered is the Indian Languages Transliteration (ITRANS) form of the respective Bengali word. Bengali is an Indo-Aryan language of India where 8.10\% \footnote{https://en.wikipedia.org/wiki/Languages\_of\_India} of the total population are $1^{st}$ language speakers and is also the official language of Bangladesh. The mother script of Bengali is the Eastern Nagari Script~\footnote{https://www.omniglot.com/writing/bengali.htm}. Our architecture utilizes fully char based sequence to sequence learning in addition to Levenshtein distance to give the final normalized form or as close to it as possible. Some additional advantages of our system is that at an intermediate stage, the back-transliterated form of the word can be fetched (i.e. word identification), which will be very useful in several cases as original tools (i.e. tools using mother script) can be utilized, for example emotion lexicons. Some other important contributions of our research are the new lexicons that have been prepared (discussed in Sec~\ref{sec3}) which can be used for building various other tools for studying Bengali-English code-mixed data.

\section{Related Work}

Normalization of text has been studied quite a lot \cite{sproat:1999normalization}, especially as it acts as a pre-processing step for several text processing systems. Using Conditional Random Fields (CRF), \citet{zhu:2007unified} performed text normalization on informal emails. \citet{dutta:2015text} created a system based on noisy channel model for text normalization which handles wordplay, contracted words and phonetic variations in code-mixed background. An unsupervised framework was presented by \citet{sridhar:2015unsupervised} for normalizing
domain-specific and informal noisy texts
using distributed representation of words. The soundex algorithm was used in \cite{sitaram:2015experiments} and \cite{sitaram:2016speech} for spelling correction of transliterated words and normalization in a speech to text scenario of code-mixed data respectively. \citet{sharma:2016shallow} build a normalization system using noisy channel framework and SILPA spell checker in order to build a shallow parser. \citet{sproat:2016rnn} build a system combining two models, where one essentially is a seq2seq model which checks the possible normalizations and the other is a language model which considers context information. \citet{jaitly:2017rnn} used a seq2seq model with attention trained at sentence level followed by error pruning using finite-state filters to build a normalization system, mainly targeted for text to speech purposes. A similar flow was adopted by \citet{zare:2017deepnorm} as well where seq2seq was used for normalization and a window of size 20 was considered for context. \citet{singh:2018automatic} exploited the fact that words and their variations share similar context in large noisy text corpora to build their normalizing model, using skip gram and clustering techniques. To the best of our knowledge, the system architecture proposed by us hasn't been tried before, especially for code-mixed data.

\section{Data Sets}
\label{sec3}

On a whole, three data sets or lexicons were created. The first data set was a parallel lexicon (PL) where the 1\textsuperscript{st} column had phonetically transliterated Bn words in Roman taken from code-mixed data prepared in \citet{mandal:2018preparing}. The 2\textsuperscript{nd} column consisted of the standard Roman transliterations (ITRANS) of the respective words. To get this, we first manually back-transliterated PL\textsubscript{col\_1} to the original word in Eastern Nagari script, and then converted it into standardized ITRANS format. The final size of the PL was 6000. The second lexicon we created was a transliteration dictionary (BN\_TRANS) where the first column had Bengali words in Eastern Nagari script taken from samsad~\footnote{http://dsal.uchicago.edu/dictionaries/biswas-bengali/}, while the second column had the standard transliterations (ITRANS). The number of entries in the dictionary was 21850. For testing, we took the data used in \citet{mandal:2018analyzing}, language tagged it using the system described in \citet{mandal:2018language}, and then collected Bn tagged tokens. Post manual checking and discarding of misclassified tokens, the size of the list was 905. Finally, each of the words were tagged with their ITRANS using the same approach used for making PL. For PL\textsubscript{col\_1} and test data, some initial rule based normalization techniques were used. If the input string contains a digit, it was replaced by the respective phone (e.g. \textit{ek} for 1, \textit{dui} for 2, etc), and if there are \textit{n} consecutive identical characters where \textit{n} $>$ 2 (elongation), it was trimmed down to 2 consecutive characters (e.g. \textit{baaaad} will become \textit{baad}), as no word in it's standard form has more than two consecutive identical characters. 

\section{Proposed Method}
Our method is a two step modular approach comprising of two degrees of normalization. The 1$^{\circ}$ normalization module does an initial normalization and tries to convert the input string closest to the standard transliteration. The 2$^{\circ}$ normalization module takes the output from the first module and tries to match with the standard transliterations present in the dictionary (BN\_TRANS). The candidate with the closest match is returned as the final normalized string.

\section{1$^{\circ}$ Normalization Module}
The purpose of this module is to phonetically normalize the word as close to the standard transliteration as possible, to make the work of the matching module easier. To achieve this, our idea was to train a sequence to sequence model where the input sequences are user transliterated words and the target sequences are the respective ITRANS transliterations. We had specifically chosen this architecture as it has performed amazingly well in complex sequence mapping tasks like neural machine translation and summarization.

\subsection{Seq2Seq Model}
The sequence to sequence model \cite{sutskever:2014sequence} is a relatively new idea for sequence learning using neural networks. It has been especially popular since it achieved state of the art results in machine translation task. Essentially, the model takes as input a sequence \textit{X} = \{x\textsubscript{1}, x\textsubscript{2}, ..., x\textsubscript{n}\} and tries to generate another sequence \textit{Y} = \{y\textsubscript{1}, y\textsubscript{2}, ..., y\textsubscript{m}\}, where x\textsubscript{i} and y\textsubscript{i} are the input and target symbols respectively. The architecture of seq2seq model comprises of two parts, the encoder and decoder. As the input and target vectors were quite small (words), attention \cite{vaswani:2017attention} mechanism was not incorporated. 
\subsubsection{Encoder} Encoder essentially takes a variable length sequence as input and encodes it into a fixed length vector, which is suppose to summarize it's meaning taking into context as well. A recurrent neural network (RNN) cell is used to achieve this. The directional encoder reads the sequence from one end to the other (left to right in our case).
\begin{equation}
\notag
\vec{h}\textsubscript{t} = \vec{f}\textsubscript{enc}(E\textsubscript{x}(x\textsubscript{t}),\vec{h}\textsubscript{t-1})
\end{equation}
Here, \textit{E}\textsubscript{x} is the input embedding lookup table (dictionary), $\vec{f}$\textsubscript{enc} are the transfer function for the recurrent unit e.g. Vanilla, LSTM or GRU. A contiguous sequence of encodings \textit{C} = \{h\textsubscript{1}, h\textsubscript{2}, ..., h\textsubscript{n}\} is constructed which is then passed on to the decoder.
\subsubsection{Decoder} Decoder takes input context vector \textit{C} from the encoder, and computes the hidden state at time t as, 
\begin{equation}
\notag
s\textsubscript{t} =  f\textsubscript{dec}(E\textsubscript{y}(y\textsubscript{t-1}), s\textsubscript{t-1}, c\textsubscript{t})
\end{equation} 
Subsequently, a parametric function out\textsubscript{k} returns the conditional probability using the next target symbol being k. Here, the concept of teacher forcing is utilized, the strategy of feeding output of the model from a prior time-step as input.
\begin{equation}
\notag
\begin{multlined}
p(y\textsubscript{t}=k\mid y<{t}, X) = \\
\frac{1}{Z}exp(out\textsubscript{k}(E\textsubscript{y}(y\textsubscript{t}-1), s\textsubscript{t}, c\textsubscript{t}))
\end{multlined}
\end{equation}
Z is  the normalizing constant
\begin{equation}
\notag
\sum\textsubscript{j}exp(out\textsubscript{j}(E\textsubscript{y}(y\textsubscript{t}-1), s\textsubscript{t}, c\textsubscript{t}))
\end{equation} 

\subsection{Training}
\begin{figure}[h]
\centering
\includegraphics[scale=0.38]{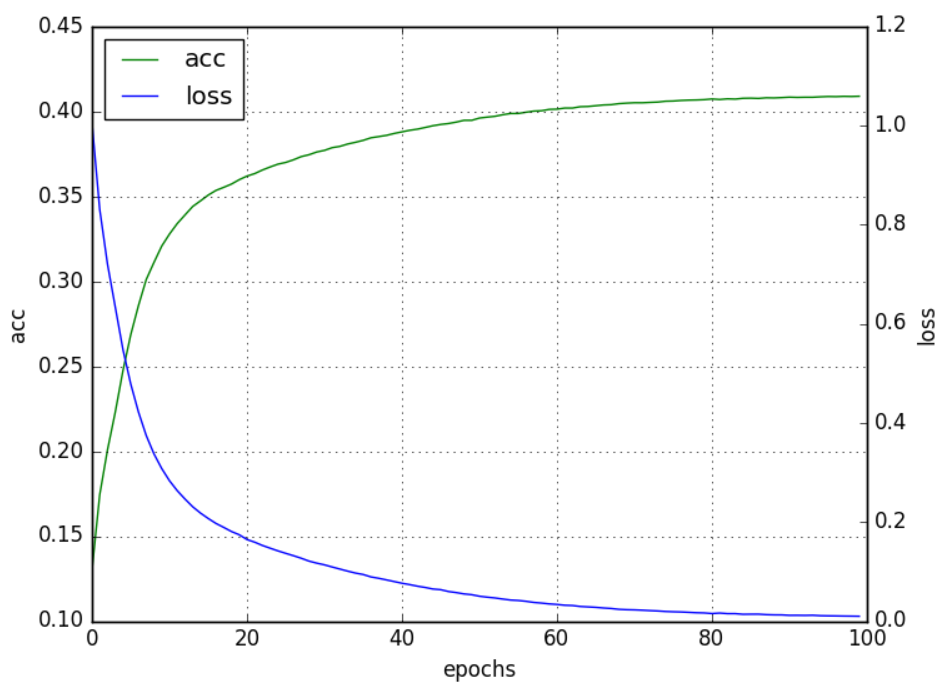}
\caption{Training accuracy and loss.}
\label{fig1}
\end{figure}

The model is trained by minimizing the negative log-likelihood. For training, we used the fully character based seq2seq model \cite{lee:2016fully} with stacked LSTM cells. The input units were user typed phonetic transliterations (PL\textsubscript{col\_1}) while the target units were respective standard transliterations (PL\textsubscript{col\_2}). Thus, the model learns to map user transliterations to standard transliterations, effectively learning to normalize phonetic variations. The lookup table \textit{E}\textsubscript{x} we used for character encoding was a dictionary where the keys were the 26 English alphabets and the values were the respective index. Encodings at character level were then padded to the length of the maximum existing word in the dataset, which was 14, and was converted to one-hot encodings prior to feeding the to the seq2seq model. We created our seq2seq model using the Keras~\cite{chollet:2015keras} library. The batch size was set to 64, and number of epochs was set to 100. The size of the latent dimension was kept at 128. Optimizer we chose was rmsprop, learning rate set at 0.001, loss was categorical crossentropy and transfer function used was softmax. Accuracy and loss graphs during training with respect to epochs are shown in Fig~\ref{fig1}.

\noindent As we can see from Fig~\ref{fig1}, the accuracy reached at the end of training was not too high (around 41.2\%) and the slope became asymptotic. This is quite understandable as the amount of training data was relatively quite low for the task, and the phonetic variations were quite high. On running this module on our testing data, an accuracy of 51.04\% was achieved. It should be noted that even a single character predicted wrongly by the softmax function reduces the accuracy.

\section{2$^{\circ}$ Normalization Module}
This module basically comprises of the string matching algorithm. For this, we have used Levenshtein distance (LD) \cite{levenshtein:1966binary}, which is a string metric for measuring difference between two sequences. It does so by calculating the minimum number of insertions, deletions and substitutions required for converting one sequence to the other. Here, the output from the previous module is compared with all the standard ITRANS entries present in BN\_TRANS and the string with the least Levenshtein distance is given as output, which is the final normalized form. If there are ties, the instance which has higher matches traversing from left to right is given more priority. Also, observing the errors from 1$^{\circ}$ normalizer, we noticed that in a lot of cases, the character pairs \{a,o\} and \{b,v\} are used interchangeably quite often, both in phonetic transliterations alone, as well as when compared with ITRANS. Thus, along with the standard approach, we tried a modified version as well where the cost of the above mentioned character pairs are same, i.e. they are treated as identical characters. This was simply done by assigning special symbols to those pairs, and replacing them in the input parameters. For example, post replacement, distance(\textit{chalo}, \textit{chala}) will become distance(\textit{ch\$l\$}, \textit{ch\$l\$}).

\section{Evaluation}
Our system was evaluated in two ways, one at word level and another at task level.

\subsection{Word Level}
Here, the basic idea was compare the normalized words with the respective standard transliterations. For this, the testing data discussed in Sec~\ref{sec3} was used. For comparison purposes, three other setups other than our proposed model (setup\_4) were tested, all of which are described in Table~\ref{table1}. 

\begin{table}[H]
\centering
\begin{tabular}{|c|c|c|c|}
\hline
\textbf{Model} & \textbf{1$^{\circ}$} \textbf{Norm} & \textbf{LD} & \textbf{Acc} \\ \hline
setup\_1 & no & standard &  58.78 \\ \hline
setup\_2 & no & modified &  61.10 \\ \hline
setup\_3 & yes & standard &  89.72 \\ \hline
setup\_4 & yes & modified &  \textbf{90.27} \\ \hline
\end{tabular}
\caption{Comparison of different setups.}
\label{table1}
\end{table}

\noindent From Table~\ref{table1}, we can see that the jump in accuracy from setup\_1 to setup\_3 is quite significant (by 30.94\%). This proves that instead of simple distance comparison with lexicon entries, a prior seq2seq normalization can have great impact on the performance. Additionally, we can also see that when modified input is given to the Levenshtein distance (LD), the accuracies achieved are slightly better. On analyzing the errors, we found out that majority (92\%) of them is due to the fact that the standard from was not present in BN\_TRANS, i.e. was out of vocabulary. These words were mostly slangs, expressions, or two words joined into a single one. The other 8\% was due to the 1$^{\circ}$ module casuing substantial deviation from normal form. For deeper analysis, we collected the ITRANS of errors due out of vocab, and on comparison with the 1$^{\circ}$ normalizations, the mean LD was calculated to be 1.89, which is suggesting that if they were present in BN\_TRANS, the normalizer would have given the correct output.

\subsection{Task Level}
For task level evaluation, we decided to go with sentiment analysis using the exact setup and data described in \citet{mandal:2018preparing}, on Bengali-English code-mixed data. All the training and testing data were normalized using our system along with the lexicons that are mentioned. Finally, the same steps were followed and the different metrics were calculated. The comparison of the systems prior (noisy) and post normalization (normalized) is shown in Table~\ref{table2}.

\begin{table}[H]
\centering
\begin{tabular}{|c|c|c|c|c|}
\hline
\textbf{Model} & \textbf{Acc} & \textbf{Prec} & \textbf{Rec} & \textbf{F1} \\ \hline
 noisy&  80.97&  81.03&  80.97&  81.20\\ \hline
 normalized&  \textbf{82.47}&  82.52&  82.47&  82.61\\ \hline
\end{tabular}
\caption{Prior and post normalization results.}
\label{table2}
\end{table}

\noindent We can see an improvement in the accuracy (by 1.5\%). On further investigation, we saw that the unigram, bigram and trigram matches with the bag of n-grams and testing data increased by 1.6\%, 0.4\% and 0.1\% respectively. The accuracy can be improved further more if back-transliteration is done and Bengali sentiment lexicons are used but that is beyond the scope of this paper.

\section{Discussion}
Though our proposed model achieved high accuracy, some drawbacks are there. Firstly is the requirement for the parallel corpus (PL) for training a seq2seq model, as manual checking and back-transliteration is quite tedious. Speed of processing in terms of words/second is not very high due to the fact that both seq2seq and Levenshthein distance calculation is computationally heavy, plus the O(n) search time. For string matching, simpler and faster methods can be tested and search area reduction algorithms (e.g. specifying search domains depending on starting character) can be tried to improve the processing speed. A simple lexical checker can be added as well before using seq2seq and/or matching module to see if the word is already in it's transliterated form.  

\section{Conclusion \& Future Work}
In this article, we have presented a novel architecture for normalization of transliterated words in code-mixed scenario. We have employed the seq2seq model with LSTM cells for initial normalization followed by evaluating Levenshthein distance to retrieve the standard transliteration from a lexicon. Our approach got an accuracy of \textbf{90.27\%} on testing data, and improved the accuracy of a pre-existing sentiment analysis system by 1.5\%. In future, we would like to collect more transliterated words and increase the data size in order to improve both PL and BN\_TRANS. Also, combining this module with a context capturing system and expanding to other Indic languages like Hindi, Tamil will be one of the goals as well.

\bibliography{emnlp2018}
\bibliographystyle{acl_natbib_nourl}

\end{document}